\documentclass[10pt, conference, compsocconf]{IEEEtran}

\usepackage{url}
\usepackage{cite}
\usepackage{graphicx} 
\usepackage{epsfig} 
\usepackage{mathptmx} 
\usepackage{times} 
\usepackage{amsmath} 
\usepackage{amssymb}  

\usepackage{gensymb}

\usepackage{epstopdf}
\usepackage[]{units}
\usepackage{color}
\usepackage{verbatim}
\usepackage{psfrag}
\usepackage{textcomp}
\usepackage{graphicx}
\usepackage{caption}
\usepackage{subcaption}
\usepackage{algorithmic}
\usepackage[Pseudocode]{algorithm}
\usepackage{bm}
\usepackage{empheq}
\usepackage{pgfplots}
\usepackage{multirow}

\begin{document}
\title{Dynamic Markers: UAV landing proof of concept}

\author{\IEEEauthorblockN{Raul Acuna, Volker Willert}
	\IEEEauthorblockA{\textit{IAT} \\
		\textit{TU Darmstadt}\\
		Darmstadt, Germany \\
		(racuna, vwillert)@rmr.tu-darmstadt.de}
}

\maketitle
\thispagestyle{empty}
\pagestyle{empty}
\begin{abstract}
In this paper, we introduce a dynamic fiducial marker which can change its appearance according to the spatiotemporal requirements of the visual perception task of a mobile robot using a camera as the sensor. We present a control scheme to dynamically change the appearance of the marker in order to increase the range of detection and to assure a better accuracy on the close range. The marker control takes into account the camera to marker distance (which influences the scale of the marker in image coordinates) to select which fiducial markers to display. Hence, we realize a tight coupling between the visual pose control of the mobile robot and the appearance of the dynamic fiducial marker. Additionally, we discuss the practical implications of time delays due to processing time and communication delays between the robot and the marker. Finally, we propose a real-time dynamic marker visual servoing control scheme for quadcopter landing and evaluate the performance on a real-world example.
\end{abstract}


\section{INTRODUCTION}

A visual fiducial marker is a known shape, usually printed in a paper which is located in the environment as a point of reference and scale for a visual task. Fiducial markers are commonly used in applications such as augmented reality, virtual reality, object tracking, and robot localization. In robotics, they are used to obtain the absolute 3D pose of a robot in world coordinates. This usually involves the distribution of several markers around the environment in known positions, or fixing a camera and detecting markers attached to the robots. This serves as a good option for ground truth in testing environments but it is not convenient for some applications due to the required environment intervention. For unknown environments, the preference is for other types of localization systems that do not rely on artificial features or a previous knowledge of the environment, i.e SLAM or Visual Odometry. Nonetheless, fiducial marker based SLAM systems are still a topic of interest~\cite{Lim2009}~\cite{Neunert2016}~\cite{Munoz-Salinas2016}, mainly in controlled environments where a ground truth is required and especially when the size of the environment is big and it is not practical to use an external localization system such as VICON.

\begin{figure}[t!]
    \centering  
   	\includegraphics[width=1.0\columnwidth]{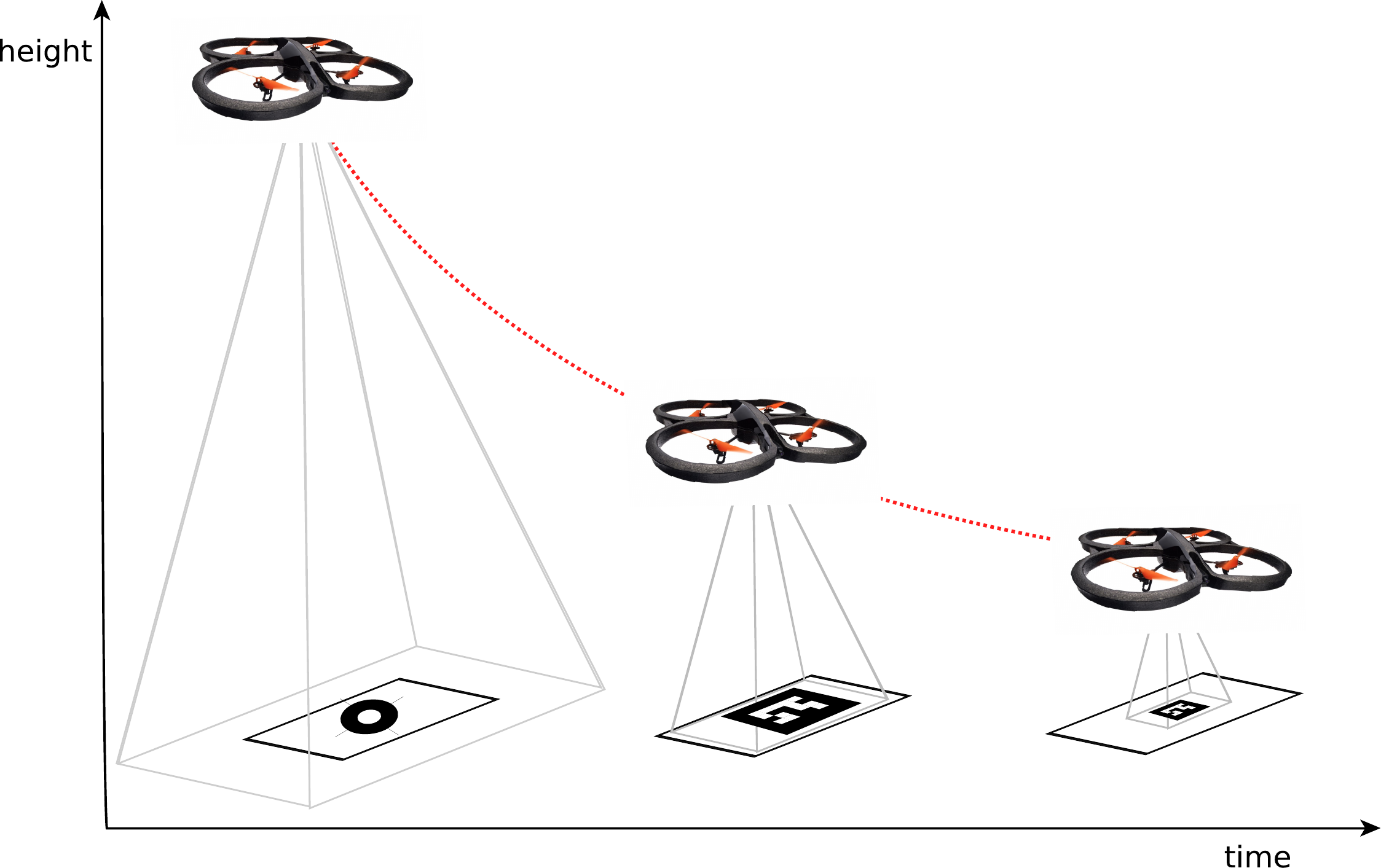}
    \caption{\small We propose a dynamic fiducial marker that can adapt over time to the requirements of the perception process. 
    This method can be integrated into common visual servoing  approaches, e.g a tracking system for autonomous quadcopter landing. 
    The size and shape of the fiducial marker can be changed dynamically to better suit the detection process dependent on 
    the relative quadcopter to marker pose. }
    
    \label{fig:dynamic_marker_quad}
\end{figure}

Fiducial markers in cooperative robotics serve as a convenient and simple inter-robot relative localization system. Since the markers are attached to the robots, no environment intervention is required. 
In the work of Howard et. al~\cite{Howard2006} fiducial markers on a team of robots are detected by the leader robot in order to combine the relative location of the other members of the team with its own global position. 
Dhiman et al.~\cite{Dhiman2013} propose a multi-robot cooperative localization system that uses reciprocal observations of camera-fiducials to increase the accuracy of the relative pose estimation. The use of top-mounted fiducial markers on UGVs is common on teams of UAV and UGV robots, these markers are then observed by the UAV. This configuration is used for coordinated navigation of a heterogeneous team of UAVs-UGVs in the work of Saska et. al~\cite{Saska2012b}, also by Mueggler et. al~\cite{Mueggler2014} for guiding a ground robot among movable obstacles using a quadcopter and more recently by Acuna et. al~\cite{Acuna2018} to perform odometry for a team of robots without external environmental features.

The vision-based autonomous landing of VTOL aerial vehicles, e.g. autonomous quadcopter landing on static or moving platforms, is an example of another field that relies on fiducial markers. The landing point is defined using a marker that can be detected by a downward looking camera in the UAV and finally, the marker can be tracked for landing by using a visual servoing controller~\cite{Saripalli2003}~ \cite{Li2011}~\cite{Lee2012a}~\cite{Bosnak2012}.

Complex fiducial markers allow the extraction of more information, for example, full 3D pose and identification of the marker between a large library of possible markers. Additionally, the number of features used for pose calculation improve the accuracy of the calculated pose. However, there is a limit on the number of features that can be present in a given market area, and this directly affects the detection distance. This means that a complex marker is harder to detect at longer distances than a simpler one, and a simple marker shape may not be able to provide full 3D pose and identification. 

The maximum range of fiducial marker detection is especially relevant for autonomous landing, for example, a large marker is wanted in order to increase the detection distance, however, if the marker is too big and the camera is close, then the marker won't be detected.

Our proposal is novel and simple: Instead of using a marker with a fixed configuration, 
we propose using a screen (such as LED/LCD displays or E-Ink displays) that can change the marker shape dynamically. A marker that changes requires a controller which must be coupled with the perception algorithm and the movement of the camera. In this paper, we present the minimal hardware/software set-up for a dynamic marker and introduce a control scheme that integrates conveniently into visual servoing. We will demonstrate that including a dynamic marker in the action-perception-cycle of the robot improves the detection range of the marker and thus improves the robot performance compared to using a static marker, which can be advantageous even considering the increase in system complexity. A great disadvantage of our proposal may be the added costs of a display screen, however in nowadays robots the presence of screens is more than common since many robots require a PC/laptop to operate, the same screens can be used for the dynamic marker.


The paper is structured as follows: In Sec.~\ref{SecII}, we introduce the basic principle of the dynamic marker and propose a visual servoing control scheme. In Sec.~\ref{SecIII}, we present the design of a dynamic marker controller design by evaluating state of the art fiducial markers for pose estimation. In Sec.~\ref{SecIV} we demonstrate our concept in a real quadcopter landing experiment and finally, we evaluate our proposal and give some conclusions.

\section{Dynamic Fiducial Marker}
\label{SecII}

The proposed concept of a dynamic marker is any kind of known feature with a configuration that can be changed as needed. Since by nature it is a separate entity from the system that performs the perception, a dynamic marker is an intelligent system that requires communication with the system that controls it.

For a minimum system configuration the following modules are needed:

\begin{enumerate}
\item A screen of any kind of display technology (LED, OLED, LCD, E-INK).
\item A basic processing unit capable of changing the image on the screen on demand.
\item A communication channel between the perception system and the display system.
\end{enumerate}

These three modules will be referred from now on as the Dynamic Marker. All these elements are commonplace nowadays. During our testing, we used a convertible laptop, an Ipad, and smartphones as dynamic markers.  It is worth noting that in previous publications, screens were used to display precise images for camera calibration~\cite{Song2008}~\cite{ZongqianZhan2008}~\cite{Ha2016}. However, to the best of our knowledge, none of those applications exploited the possibilities of performing dynamic changes to the image based on the feedback from the perception task. It is precisely this feedback what makes a dynamic marker an interesting concept for control applications.

\subsection{Pose based visual servoing}

Traditional monocular visual servo control uses the image information captured by the camera to control the movement of a robotic platform. It can be separated into two categories, Image-Based Visual Servoing (IBVS) which is based directly on the geometric control of the image features, and Position Based Visual Servoing (PBVS) which projects known structures to estimate a pose which is in turn used for robot control. We are going to focus on PBVS for our dynamic marker analysis but the same concepts apply to IBVS~\cite{Francois2006}.

The goal in a visual servoing approach is to minimize an error $\mathbf{e}(t)$ defined by

\begin{equation}
\mathbf{e}(t) = \mathbf{s}(\mathbf{m}(t), \mathbf{a}) - s^*
\label{eq:error}
\end{equation}

The image measurements $\mathbf{m}(t)$ are used to calculate a vector of visual features $\mathbf{s}(\mathbf{m}(t), \mathbf{a})$. For PBVS, $\mathbf{s}$ is defined in terms of the parametrization $\mathbf{a}$ used to obtain the camera pose which includes the camera intrinsic parameters and the 3D model of the object (in our case the fiducial marker). We are going to maintain the visual servoing frame conventions, the current camera frame $\mathit{F}_c$, the desired camera frame $\mathit{F}_c^*$ and the marker reference frame $\mathit{F}_m$. A translation vector $^c\mathbf{t}_m$ gives the coordinate of the marker frame relative to camera frame and the coordinate vector $^{c^*}\mathbf{t}_c$ gives the coordinate of the current camera frame relative to the desired camera frame. The matrix $\mathbf{R} = ^{c^*}\mathbf{R}_c$ is the rotation matrix that defines the orientation from current camera frame to the desired frame, and $\theta\mathbf{u}$ is the angle-axis representation of the rotation.

We are going to define $\mathbf{t}$ in relation to the desired camera frame $\mathit{F}_c^*$, then $\mathbf{s}= (^{c^*}\mathbf{t}_c, \theta\mathbf{u})$, $\mathbf{s}^*=0$ and $\mathbf{e}=s$. A common control approach to minimize the error is to control the velocity of the camera. The rotational and translational motion can be decoupled in this case and the control scheme will be:

\begin{empheq}[left = \empheqlbrace]{align}
	\bm{v}_c &= -\lambda\mathbf{R}^{\top}\ {^{c^*}\mathbf{t}_c}\\
    \bm{w}_c &= -\lambda\theta\mathbf{u}
\end{empheq}

Where $\bm{v}_c$ and $\bm{w}_c$ are the camera translational and rotational velocities. A PBVS approach is very similar to traditional pose robot control. By using the previously defined controller, it is possible to control separately the translational and rotational velocities of the camera to converge to the desired set of features.


\subsection{Dynamic marker controller}
\begin{figure}[t!]
    \centering  
   	\includegraphics[width=0.8\columnwidth]{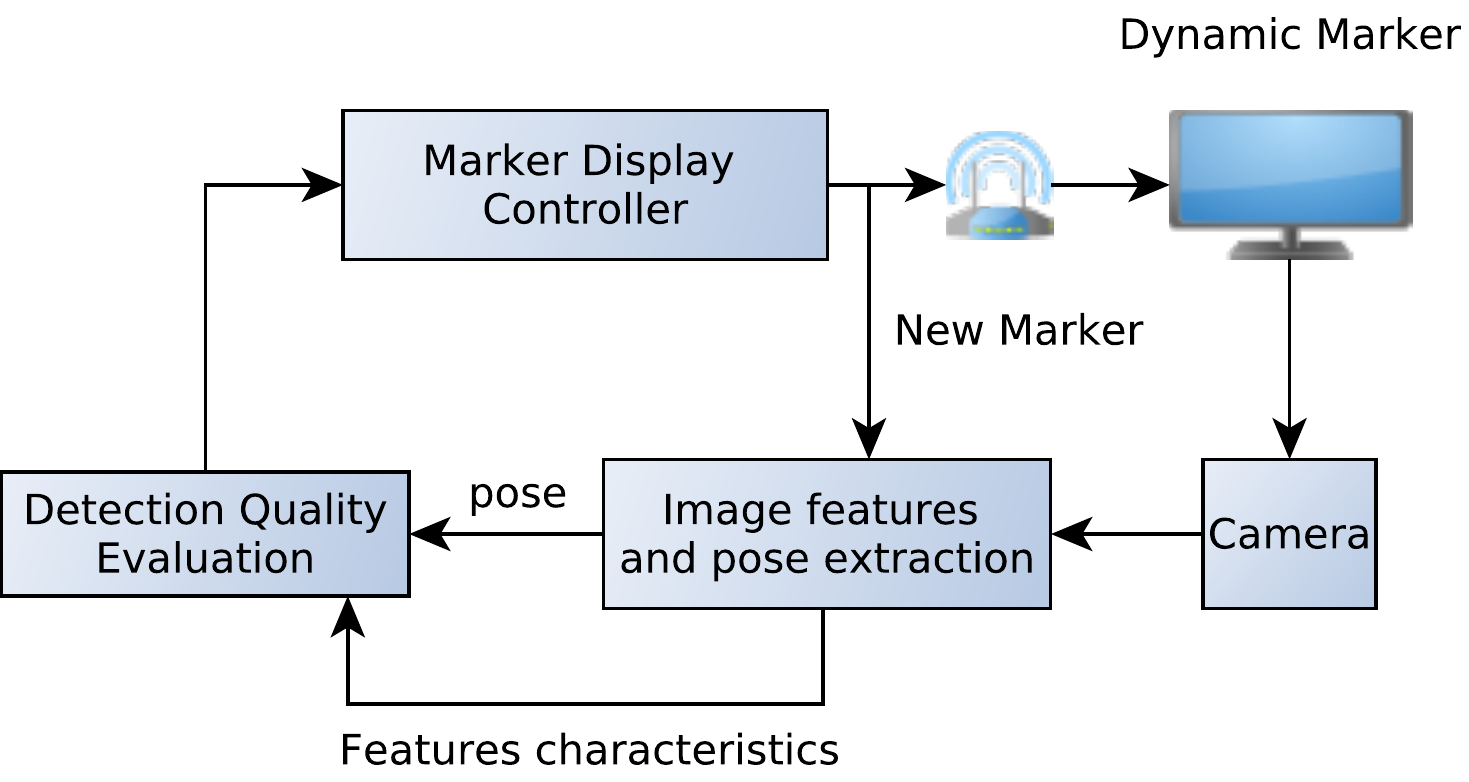}
    \caption{\small Dynamic marker control diagram. }
	\vspace{-0.50cm}    
    \label{fig:marker_control}
\end{figure}

A dynamic marker is a fiducial marker that can be controlled in order to optimize the image features extraction process and posterior pose estimation.
There are two control objectives for the dynamic marker control: 1) The marker size should be controlled in such a way that it keeps being in the field of view  of the robot allowing for small robot pose changes within some bounds, and
2) the marker type and appearance should be selected so that it increases the accuracy of the pose estimation for the current pose. For example, to increase the range, a simple marker may be used for the long range and a more complex one (with full 6DOF pose capabilities) can be displayed in the close range with varying scales. 

The proposed control loop for a dynamic marker is presented in Fig.~\ref{fig:marker_control}. For the initial analysis, we assume that both the camera and the dynamic marker are static, with a relative pose from a marker frame to camera frame defined by $^c\mathbf{T}_m$. 
It is assumed that for a given camera state (pose and dynamics) and given camera intrinsic parameters, there should exist an ideal marker configuration which optimizes the pose calculation. We use this premise as the basis for our controller design. 

The vector $\mathbf{s}$, in this case, represents the currently estimated features, and the vector $\mathbf{s}^*$ contains the optimal set of features for the current state. Analogous to the visual servoing approach, the goal of the control is to minimize the error $\mathbf{e}$ defined as in \eqref{eq:error}. The error in PBVS is minimized by moving the camera, in dynamic marker control we additionally change the appearance of the marker to minimize the error. 

After each marker detection and subsequent pose estimation, an evaluation of the current state of the detection and estimation is performed, which may be defined as a function of all or some of the following: 
1) the calculated pose,
2) the vector of image features,
3) the camera intrinsic parameters and 
4) the overall image quality (e.g noise, contrast, brightness, blur). 
Based on this evaluation, the controller selects the marker shape and scale to a configuration that can increase the performance of the system.

Now,  the vector $\mathbf{s}$ depends both on the image measurements $\mathbf{m}(t)$ and the set of parameters $\mathbf{a}$, and $\mathbf{a}$ represents our knowledge about the system including the camera intrinsic parameters plus the 3D model of the marker. If the marker 3D configuration changes it is also necessary to update $\mathbf{a}$ in the marker recognition algorithm, this means that for a dynamic marker the $\mathbf{a}$ parameter changes over time so it will be represented as $\mathbf{a}(t)$.  This update path can be observed in the control diagram of Fig.~\ref{fig:marker_control}.

Finally, since the dynamic marker in itself is a separate system, 
the new marker configuration has to be sent through a communication channel 
(e.g. Wifi) so the screen can be updated with the new image. 
Ideally there must be a confirmation sent back from the dynamic marker stating that the new marker was in fact updated.

\subsection{Dynamic marker PBVS}
\begin{figure}[t!]
    \centering  
    \vspace{0.20cm} 
    \includegraphics[width=0.8\columnwidth]{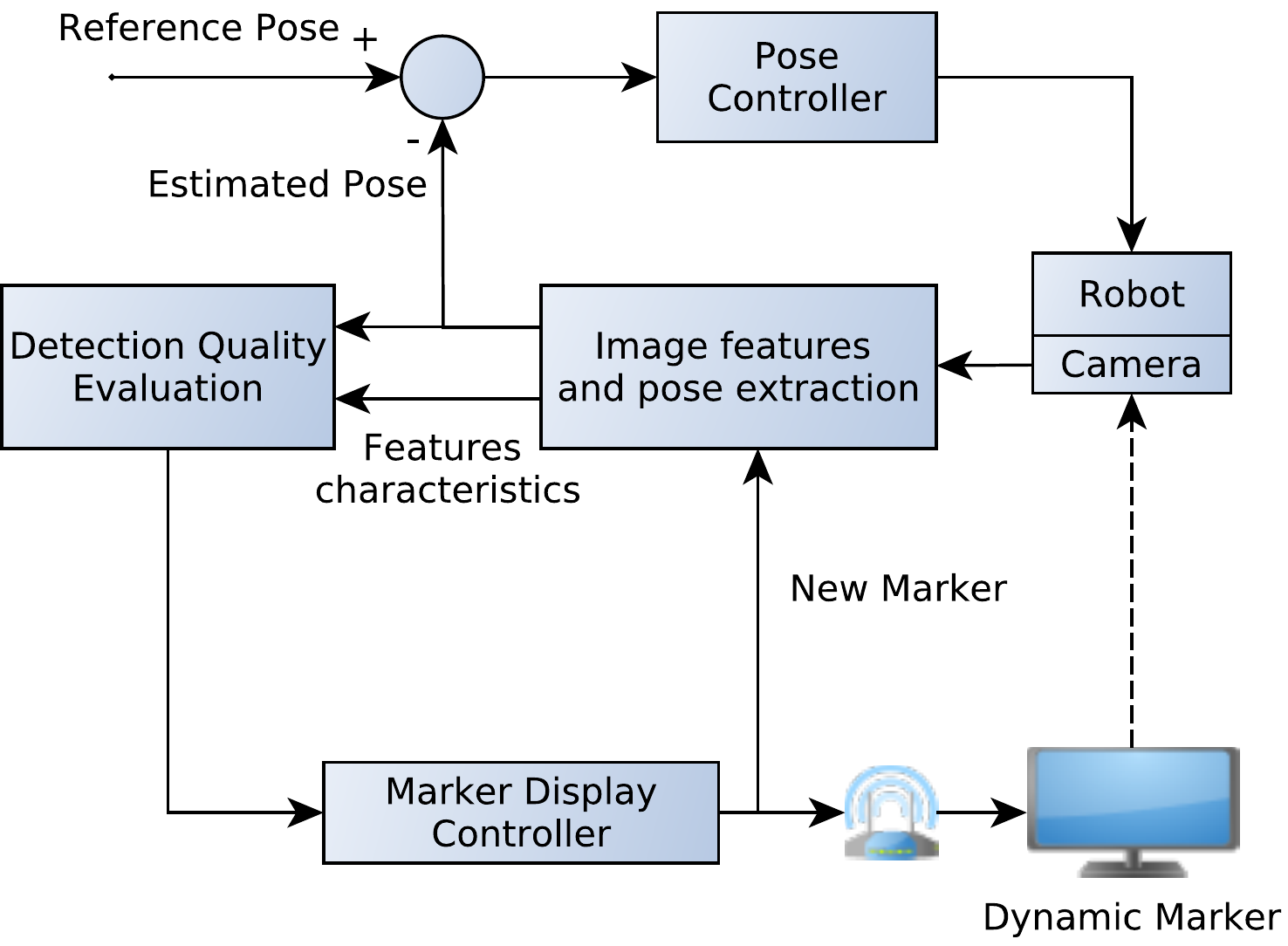}
    \caption{\small A dynamic marker used with a virtual servoing control approach. }
    \vspace{-0.20cm} 
    \label{fig:dynamic_marker_PBVS}
\end{figure}

Now that the fundamentals of the dynamic marker control are defined, we can integrate this system into a PBVS approach.  For this, we assume that the camera is part of a robotic platform and that its movement can be controlled. Fig.~\ref{fig:dynamic_marker_PBVS} shows the proposed control diagram. It is possible to define two separate control loops, the top one related to camera movement and the bottom one related to marker 3D model changes, both of them try to minimize the overall pose error simultaneously. The dynamic marker tries to maximize the marker detection and pose accuracy, which in turn results in better pose estimates for the PBVS. 

There is an interesting consequence of having a dynamic marker in a PBVS control loop. If the marker scale and orientation are changed without updating $\mathbf{a}(t)$, it is possible to directly control the pose of the robotic platform by only changing the marker. This behavior will be shown in the experiments.

\subsection{System delays analysis}

\begin{figure}[t!]
    \centering  
    \vspace{0.20cm} 
   	\includegraphics[width=\columnwidth]{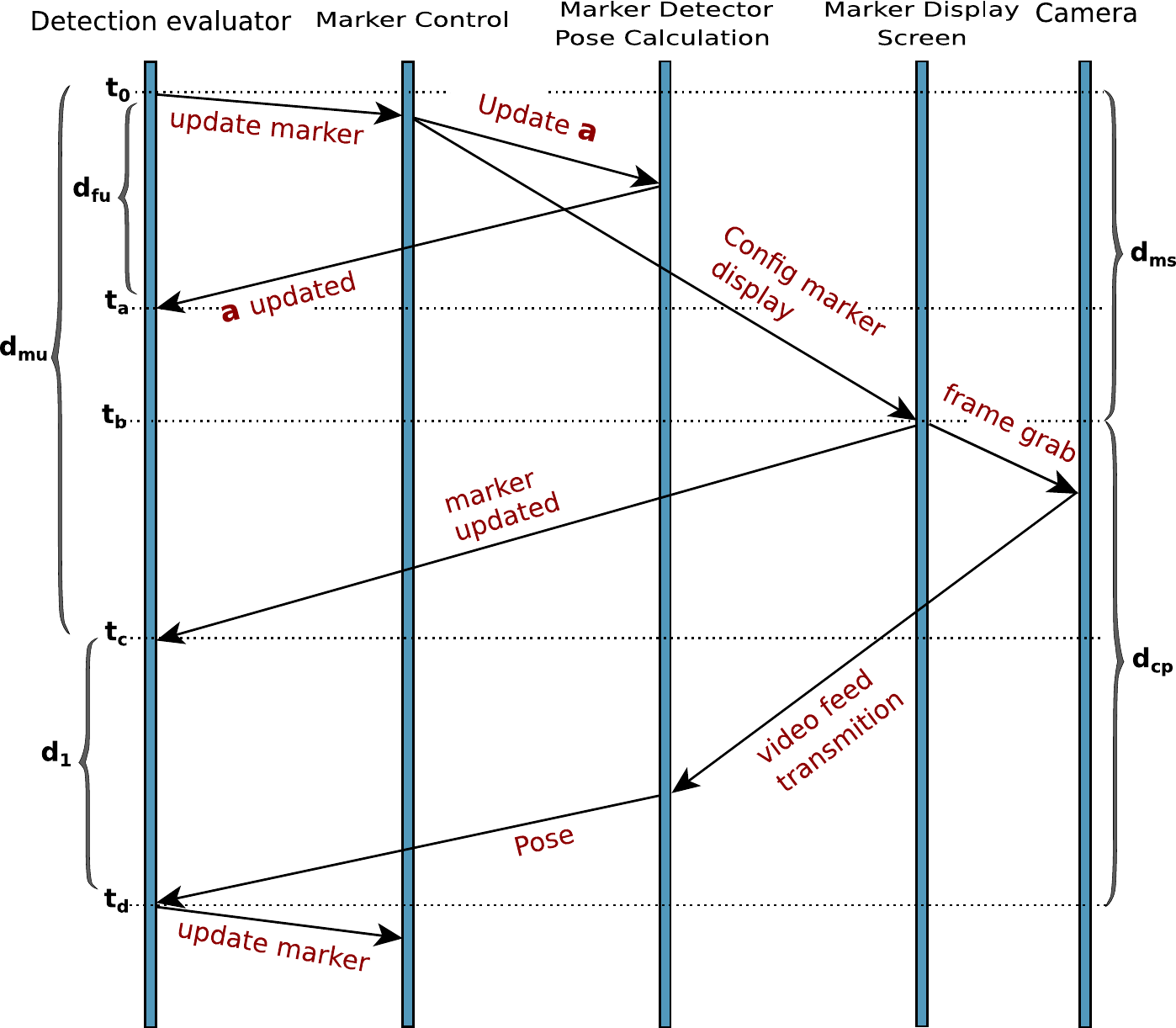}
    \caption{\small Detail of the timing of each event during the dynamic marker PBVS control loop. }
    \vspace{-0.20cm}     
    \label{fig:delays}
\end{figure}

There is a race condition on the control loop that must be considered since a proper synchronization between the feature detection algorithm and the marker display is required. If the marker is changed on the display, but the feature detection algorithm is not updated with the new value of $\mathbf{a}(t)$ before the updated image comes from the camera, then a wrong pose will be calculated. In Fig.~\ref{fig:delays} a complete timeline of the important events during the control loop is presented. This diagram will be used to precisely point out the important delays and how to tackle the race conditions. 

The relevant delays on the system are:

\subsubsection{Feature detector updated confirmation $d_{fu}$} Time needed to send the new marker 3D model parameters $\mathbf{a}(t)$ 
to the feature detector and receive a confirmation of the successful change: $d_{fu} = t_a-t_0$.

\subsubsection{Marker on screen $d_{ms}$} Time required to send the new marker command to the screen 
and for it to be displayed on the screen: $d_{ms} = t_b - t_0$.

\subsubsection{Marker updated confirmation $d_{mu}$} Time that passes since the instant a new marker command is transmitted to the dynamic marker and a confirmation is received. 
Note that in between these two instants there is the instant $t_b$ in which the new marker is actually being displayed. This delay can change in a non-deterministic way depending on the type of communication. In practice it is difficult, if not impossible, to know exactly when the new marker is on the screen. However, $d_{mu}$ will be used as an upper bound: $d_{mu} = t_c - t_0$.

\subsubsection{Capture and pose $d_{cp}$} Time passed between $t_b$ (a new marker is on screen) and $t_d$ 
(a pose calculation is ready to be used). Two critical delays play a role here, 
first the frame grab delay $d_{frame}$, given by the frame rate of the camera, and second, 
the delay on video transmission $d_{video}$ which can be considerably longer in Wifi or RF transmissions. 
The pose estimation $d_{pose}$ is faster so it doesn't play a relevant role: 
$d_{cp} = t_d - t_b = d_{frame} + d_{video} + d_{pose}$.

\subsubsection{Marker capture loop $d_{cl}$} Time passed between $t_0$ when a new marker command is transmitted 
to the dynamic marker and $t_d$ (a pose estimation is ready to be used): $d_{cl} = d_{ms} + d_{cp}$.

\subsubsection{$d_{diff}$} Is the time difference between a complete marker capture loop $d_{cl}$ 
and a marker updated confirmation $d_{mu}$. Depending on the speed of the camera and the video transmission 
it is possible that $d_{cl}$ takes less time than $d_{mu}$.

We are going to define $d_{wait}$ as the amount of time that has to pass between valid poses ensuring 
that the feature detector is in fact calculating a pose based on the 
parameters of the currently displayed marker and not the previous one. A safe choice for $d_{wait}$ is
\begin{equation}
d_{wait} = max(d_{cl},d_{mu}) 
\label{eq:delay_wait}
\end{equation}

This means, that at each new marker configuration loop, the feature detector must wait for $d_{wait}$ milliseconds to provide new valid pose estimates.

However, this represents only an absolute maximum. It is possible to make an optimization if the following conditions are true: 
1) $d_{mu} < d_{cl}$, 2) $d_{fu}$ is relatively small, 3) $d_{cp}$ is basically constant time. This means, 
that the only non-deterministic delay on the system is $d_{mu}$. 
To take advantage of this knowledge we can move the process of updating the parameter $\mathbf{a}$ in the feature detector 
(the portion corresponding to $d_{fu}$ in Fig.~\ref{fig:delays}) to the final part of the loop in any instant after $t_c$ and close to $t_d$. 
This would mean that between $t_0$ and $t_d-d_{fu}$ the feature detector will calculate valid poses based on images of the old marker. 
By doing this, the waiting time is reduced to $d_{wait} = d_{fu} + d_{safety}$, where $d_{safety}$ is a small value to compensate small deviations 
on the video capture and pose calculation process. In our tests, we defined $d_{safety} = d_{frame}$ with good results. This means, 
that for each marker update, we only invalidate the measurement of one frame. Of course, this is highly dependent on the hardware configuration.


\section{Dynamic Marker PBVS controller design for autonomous quadcopter landing problem}
\label{SecIII}

To validate our proposal, the design of a dynamic marker PBVS controller will be presented in this section. We have chosen the landing of an autonomous quadcopter as a test-bed since it presents the typical problems related to PBVS on a dynamic platform. For the design of the dynamic marker controller, it is necessary to characterize the perception problem with regular fiducial markers in order to understand which dynamic changes are needed.

\subsection{Fiducial markers for landing}


The major problem for fiducial marker quadcopter landing is the detection range of the marker. It is preferred to have long detection distance but also full pose information and marker identification is required at some point. The final centimeters of the landing are also critical. This means, that the marker should be observable at short range. 
This set of requirements presents a problem when choosing a marker, and usually, a trade-off is done. Thus, the design of the controller will be based on two requirements. First, the ability to display markers from different marker families and second, to scale the marker based on the camera-to-marker distance.


%

Two kinds of marker families were selected, one with high complexity full 6DOF but low accuracy at long distances (Aruco~\cite{aruco}) and another with low complexity longer range but without yaw angle estimation capabilities (Whycon~\cite{Krajnik2014}). As a high complexity marker, Aruco is a convenient choice, since it is now part of OpenCV and there are several implementations on ROS. For a similar reason, the low complexity marker will be Whycon, this marker has been successfully used in several robotic applications due to its accuracy, simplicity and low processing time and also has a convenient ROS implementation. However, it is not capable of providing yaw angle information or a library of different markers. From the Aruco and Whycon papers plus our own tests, we arrived at the following conclusions: The accuracy of Aruco greatly decreases with the distance to the camera, and surprisingly the Whycon accuracy remains almost constant. Whycon is a good alternative for all ranges (for position estimation only), while Aruco is only good at small marker-to-camera distances. With our camera, the maximum detection distance for Aruco was $4.4 \ m$ and for Whycon $13.181 \ m$. Nonetheless, yaw angle estimation is required for quadcopter heading aligning to the landing marker, so Whycon cannot be used at all times. Regarding the rotation estimation results, it is observed that Whycon presents in some cases completely wrong rotation estimates and in other cases correct values but with the wrong sign. In contrast, Aruco performs better for rotation estimates. We also made comparisons between the markers printed on paper and the markers displayed on the screen and we did not find any significant difference, besides reflection problems in the LCD screen, which is, of course, a disadvantage, reflections are not present in E-Ink screens though but they are slower to refresh.

From these findings, it is defined that the dynamic marker would use Whycon as a long-range position only marker and Aruco as close range full pose marker. The limits of the transition have to be defined depending on the camera that is going to be used and the size of the display screen. 

\subsubsection{Scale change based on the FOV}
\begin{figure}[ht]
    \centering  
    \vspace{0.20cm}
   	\includegraphics[width=0.38\textwidth]{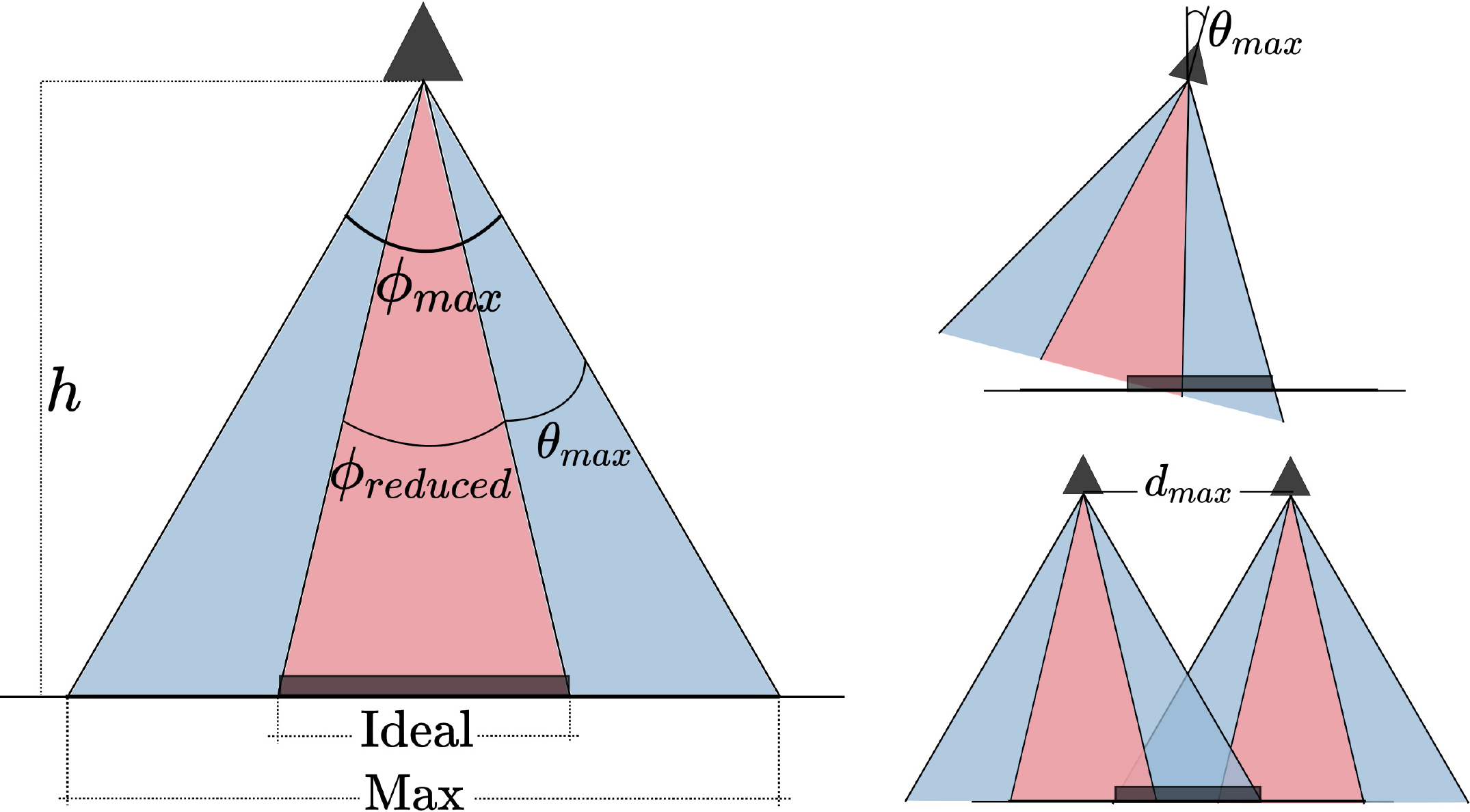}
    \caption{\small An ideal marker should have a size that allows a successful detection while leaving some extra room in the field of view for the movement of the camera. }\label{fig:fov}
    \vspace{-0.30cm}
\end{figure}

For automatic scaling of the marker, we propose a scaling rule based on the camera field of view. The desired marker size $m_{size}$ at a given marker-to-camera distance $h$ has to fit inside a reduced field of view of the camera $\phi_{reduced}$. This reduced field of view allows some room for the movement of the UAV. The marker size will be defined as a function of the following form: 
$m_{size} = f(\phi_{max},h,s)$, were $\phi_{max}$ is the maximum camera vision angle which can be obtained from the camera intrinsic parameters, 
$h$ is the marker-to-camera distance and $s$ is the scaling factor. Fig.~\ref{fig:fov}  represents the reduced field of view with the ideal and maximum marker sizes.

By using simple geometrical calculation, the following equation for the optimal marker size for a given $h$ can be found:
\begin{equation}
\label{eq:marker_size}
m_{size} = 2*h*tan(\phi_{max}*s)\, ,
\end{equation}

were $0<s\leq 1$. If $s=1$, then the size of the marker is the maximum for that given field of view (Fig.~\ref{fig:fov}).


The minimum value for $s$ depends on the minimum amount of pixels required for the marker identification algorithm. Choosing a value for $s$ can also be seen as choosing an angle $\theta_{max}$ of camera freedom. This angle can be calculated by: $\theta_{max} = \phi_{max}/2 - atan(m_{size}/(2*h)$.

\subsubsection{Final dynamic marker controller design}
Now it is possible to define the dynamic marker controller for quadcopter landing. Two marker families were selected, Aruco for low range and Whycon for the rest. Aruco will be scaled according to \eqref{eq:marker_size}. Another feature of Aruco is also exploited, the board of markers. When the size of the Aruco marker is small, the rest of the screen will be filled with more Aruco markers with different IDs, all of them form part of the same coordinate system which increases accuracy on close ranges. At the start of the system, the initial marker will be Whycon to ensure detection.

\section{Autonomous Quadcopter Landing using dynamic fiducial marker}
\label{SecIV}

Our experimental setup for quadcopter landing consists in an AR.Drone Parrot 2.0 quadcopter with a custom wireless camera and landing legs. 
All the image processing and control is done in a ground station that sends the commands back to the quadcopter through Wifi using the ROS 
ardrone autonomy package. We implemented an observer and a predictor module to cope with the Wifi delay problems of the AR.Drone and a velocity 
controller based on these predictions. A foldable laptop with a 13.1~inch OLED screen was selected as the dynamic marker. 
The code for the dynamic marker was implemented in Openframeworks using Websockets and connected to ROS via ros-bridge. 
For marker recognition the $ar\_sys$ and $whycon\_ros$ packages were used for Aruco and Whycon detection, respectively, 
with some modifications to the $ar\_sys$ package for dynamic marker reconfiguration. 
On top, we have a simple PBVS controller based on the velocity controller that we developed for the AR.Drone. 
For this camera/display configuration the maximum size of the display marker is $15 \ cm$ and the maximum range for Aruco is $1.5 \ m$, 
so the switching point between Aruco and Whycon was defined at $1.2 \ m$.

\begin{figure*}[ht]
	\centering
	\vspace{0.20cm}
	\includegraphics[width=1\textwidth]{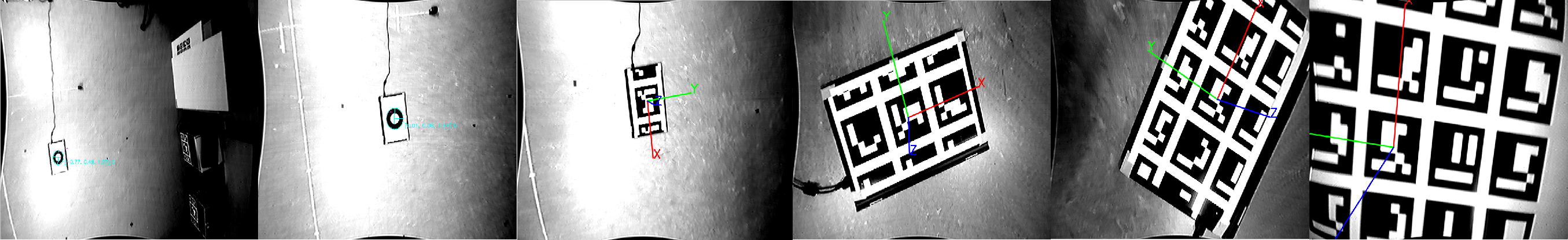}
	\caption{\small Camera frames during the landing procedure. 
		Notice the change from Whycon to Aruco in the third frame and the start of the yaw rotation correction. 
		In the frames 4, 5 and 6 it is possible to see the dynamic change of the scale.}
	\vspace{+0.00cm}
	\label{fig:camera_frames}
\end{figure*}

\subsection{Experiment, landing with a dynamic marker} 
\begin{figure}[t!]
    \centering  
    \vspace{0.20cm}
    \includegraphics[width=0.5\textwidth]{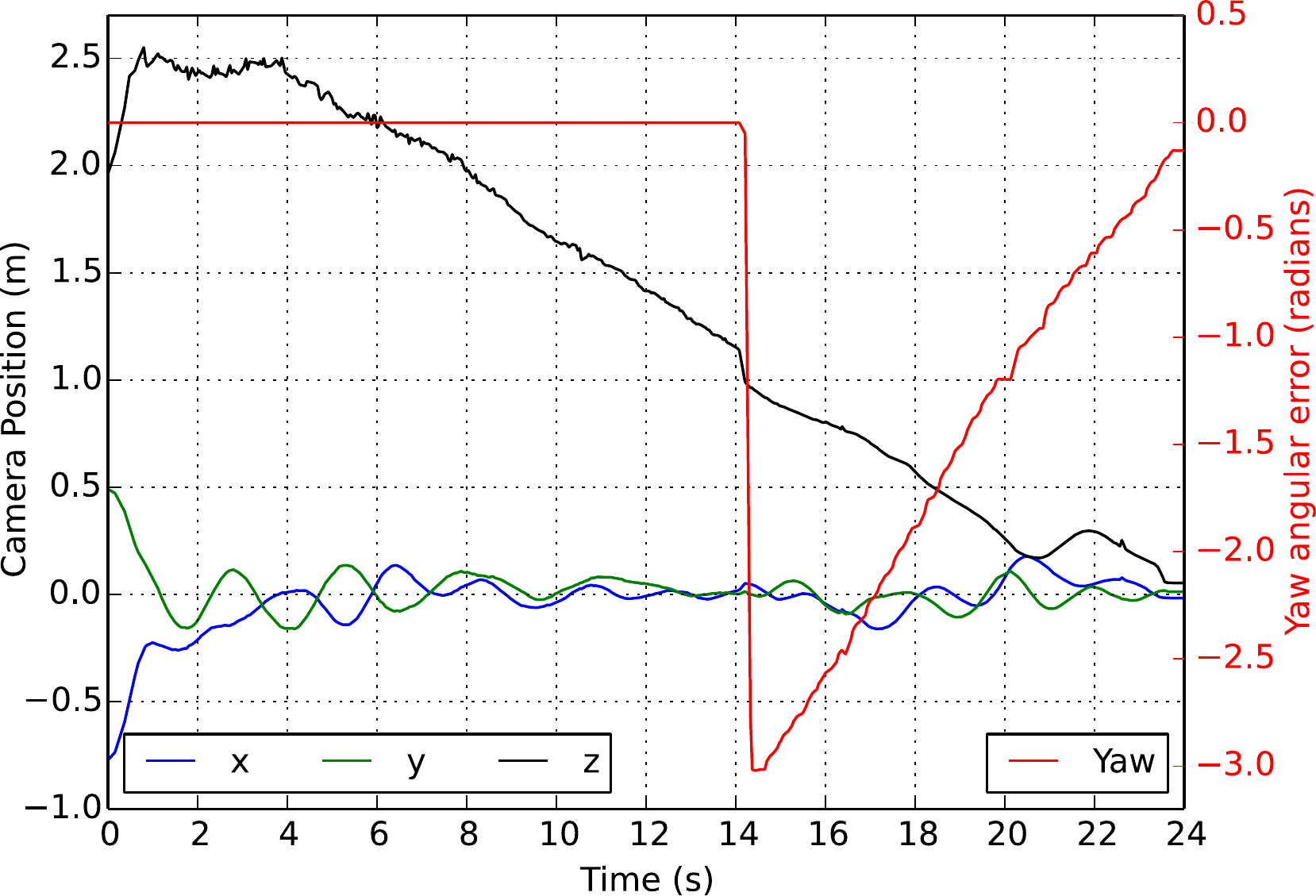}        		
    \caption{\small A successful landing using a dynamic marker. From $t=0$ to $t=14$ the displayed marker was Wycon, then Aruco board with a dynamic change of scale. Notice the yaw angle correction as soon as Aruco is detected. }\label{fig:landing_using_ref}
    \vspace{-0.40cm}
\end{figure}

The quadcopter was flown manually to a height greater than $2 \ m$ to a position where the dynamic marker was in the field of view. The PBVS was activated to track the marker at a height of $2.5 \ m$ and finally, the landing signal was sent.
The quadcopter then descended at a constant speed until the final landing was performed. 
The result of one of the typical landings can be seen in Fig.~\ref{fig:landing_using_ref}. 
Notice how the yaw angle error is corrected as soon as the dynamic marker changes into Aruco at $t=14 \ s$. 
The landing is performed smoothly and the final error for this test was $3.5 \ cm$ from the center of the marker. 
Fig.~\ref{fig:camera_frames} shows how the display changes according to dynamic marker design. 
Extensive testing was performed with this setup with more than 50 successful landings, with an average error of $4,8 \ cm$. 
In comparison if a static marker is used, either the detection range is limited, not possible to land from the configured height (by choosing Aruco) or it is impossible to align the quadcopter with the landing platform (when choosing only Whycon), proving the advantages of a dynamic marker for visual servoing based landing.

%

The PBVS and the marker are tightly coupled in a dynamic marker. Both systems are intertwined if one changes the marker without updating the $\mathbf{a}(t)$ parameter. For example, if the size of the marker is reduced by half without updating $\mathbf{a}$, then the pose estimator will calculate a "virtual" height that is twice as high as the real one, and the platform will move down to compensate. This can be used to control the platform in a unilateral way by only changing the marker, e.g. the heading of the quadcopter may be controlled by rotating the dynamic marker.



\section{Conclusions and future work}
\label{SecV}

We presented a novel concept of a dynamic fiducial marker integrated into a visual servoing control approach for UAV landing. A control architecture inspired by PBVS with timing analysis was also introduced. Our proof-of-concept system uses two different fiducial marker families to prove that a minimal system can be built to couple action and perception for UAV landing in real-time. However, this a very specific application, and the discrete selection from different marker families based on the range is only a basic objective function.

From these first results, there are many more interesting problems to study. It should be possible to obtain explicit objective functions to define an optimal marker shape for a given camera to marker pose. The definition of these objective functions and the development of more dynamic shapes (not based on already available fiducial markers) are the focus of future research. The coupling between the dynamic marker and the PBVS is an interesting research area since the dynamics of the marker can be changed instantly in time and thus is much faster than the visual servoing control of the mobile robot. The design of the marker could be extended to fully take advantage of the temporal domain, e.g. showing marker codes for identification in a temporal sequence. The number of features and their configuration could be dynamically optimized to improve pose estimation accuracy.
Finally, a more formal comparison of static traditional fiducials vs the dynamic marker is needed, by using, for example, an external camera system (e.g. Optitrack) for different camera poses since there are potential angle restrictions on LCD screens due to reflections. 







{\small
\bibliography{IEEEabrv,bib_iros2017}

\begin{thebibliography}{10}
\providecommand{\url}[1]{#1}
\csname url@samestyle\endcsname
\providecommand{\newblock}{\relax}
\providecommand{\bibinfo}[2]{#2}
\providecommand{\BIBentrySTDinterwordspacing}{\spaceskip=0pt\relax}
\providecommand{\BIBentryALTinterwordstretchfactor}{4}
\providecommand{\BIBentryALTinterwordspacing}{\spaceskip=\fontdimen2\font plus
\BIBentryALTinterwordstretchfactor\fontdimen3\font minus
  \fontdimen4\font\relax}
\providecommand{\BIBforeignlanguage}[2]{{%
\expandafter\ifx\csname l@#1\endcsname\relax
\typeout{** WARNING: IEEEtran.bst: No hyphenation pattern has been}%
\typeout{** loaded for the language `#1'. Using the pattern for}%
\typeout{** the default language instead.}%
\else
\language=\csname l@#1\endcsname
\fi
#2}}
\providecommand{\BIBdecl}{\relax}
\BIBdecl

\bibitem{Lim2009}
H.~Lim and Y.~S. Lee, ``{Real-Time Single Camera SLAM Using Fiducial
  Markers},'' in \emph{ICCAS-SICE}, 2009, pp. 177--182.

\bibitem{Neunert2016}
M.~Neunert, M.~Bloesch, and J.~Buchli, ``{An Open Source, Fiducial Based,
  Visual-Inertial Motion Capture System},'' in \emph{Int. Conf. Inf. Fusion},
  2016.

\bibitem{Munoz-Salinas2016}
R.~Mu{\~{n}}oz-Salinas, M.~J. Mar{\'{i}}n-Jimenez, E.~Yeguas-Bolivar, and
  R.~Medina-Carnicer, ``{Mapping and Localization from Planar Markers},''
  \emph{CoRR}, vol. abs/1606.0, pp. 1--14, 2016.

\bibitem{Howard2006}
a.~Howard, ``{Experiments with a Large Heterogeneous Mobile Robot Team:
  Exploration, Mapping, Deployment and Detection},'' \emph{Int. J. Rob. Res.},
  vol.~25, no. 5-6, pp. 431--447, 2006.

\bibitem{Dhiman2013}
V.~Dhiman, J.~Ryde, and J.~J. Corso, ``{Mutual localization: Two camera
  relative 6-DOF pose estimation from reciprocal fiducial observation},''
  \emph{IEEE Int. Conf. Intell. Robot. Syst.}, pp. 1347--1354, 2013.

\bibitem{Saska2012b}
M.~Saska, V.~Vonasek, T.~Krajnik, and L.~Preucil, ``{Coordination and
  navigation of heterogeneous UAVs-UGVs teams localized by a hawk-eye
  approach},'' \emph{Int. J. Rob. Res.}, vol.~33, pp. 1393--1412, 2014.

\bibitem{Mueggler2014}
E.~Mueggler, M.~Faessler, F.~Fontana, and D.~Scaramuzza, ``{Aerial-guided
  Navigation of a Ground Robot among Movable Obstacles},'' in \emph{Proc. IEEE
  Int. Symp. Safety, Secur. Rescue Robot.}, 2014.

\bibitem{Acuna2018}
R.~Acuna, Z.~Li, and V.~Willert, ``Moma: Visual mobile marker odometry,''
  \emph{International Conference on Indoor Positioning and Indoor Navigation
  (IPIN)}, 2018.

\bibitem{Saripalli2003}
S.~Saripalli, J.~Montgomery, and G.~Sukhatme, ``{Visually guided landing of an
  unmanned aerial vehicle},'' \emph{IEEE Trans. Robot. Autom.}, vol.~19, no.~3,
  pp. 371--380, 2003.

\bibitem{Li2011}
W.~Li, T.~Zhang, and K.~K{\"{u}}hnlenz, ``{A vision-guided autonomous quadrotor
  in an air-ground multi-robot system},'' in \emph{Proc. IEEE Int. Conf. Robot.
  Autom.}\hskip 1em plus 0.5em minus 0.4em\relax Shanghai: IEEE, 2011, pp.
  2980--2985.

\bibitem{Lee2012a}
D.~Lee, T.~Ryan, and H.~J. Kim, ``{Autonomous landing of a VTOL UAV on a moving
  platform using image-based visual servoing},'' \emph{Proc. IEEE Int. Conf.
  Robot. Autom.}, pp. 971--976, 2012.

\bibitem{Bosnak2012}
M.~Bo{\v{s}}nak, D.~Matko, and S.~Bla{\v{z}}i{\v{c}}, ``{Quadrocopter hovering
  using position-estimation information from inertial sensors and a high-delay
  video system},'' \emph{J. Intell. Robot. Syst. Theory Appl.}, vol.~67, no.~1,
  pp. 43--60, 2012.

\bibitem{Song2008}
Z.~Song and R.~Chung, ``{Use of LCD panel for calibrating
  structured-light-based range sensing system},'' \emph{IEEE Transactions on
  Instrumentation and Measurement}, vol.~57, no.~11, pp. 2623--2630, 2008.

\bibitem{ZongqianZhan2008}
{Zongqian Zhan}, ``{Camera calibration based on liquid crystal display
  (lcd)},'' \emph{Isprs}, no. LCD, 2008.

\bibitem{Ha2016}
H.~Ha, Y.~Bok, K.~Joo, J.~Jung, and I.~S. Kweon, ``{Accurate camera calibration
  robust to defocus using a smartphone},'' \emph{Proc. of the IEEE Int. Conf.
  on Computer Vision}, vol. 11-18-Dece, no. 2011, pp. 828--836, 2016.

\bibitem{Francois2006}
C.~Francois and S.~Hutchinson, ``{Visual Servo Control Part I : Basic
  Approaches},'' \emph{IEEE Robot. Autom. Mag.}, vol.~13, no. December, pp.
  82--90, 2006.

\bibitem{aruco}
S.~Garrido-Jurado, ``{Automatic generation and detection of highly reliable
  fiducial markers under occlusion},'' \emph{Pattern Recognit.}, vol.~4, no.~6,
  pp. 2280--2298, 2014.

\bibitem{Krajnik2014}
T.~Krajn{\'{i}}k, M.~Nitsche, J.~Faigl, P.~Vanek, M.~Saska, L.~Preucil,
  T.~Duckett, and M.~Mejail, ``{A Practical Multirobot Localization System},''
  \emph{J. Intell. Robot. Syst. Theory Appl.}, vol.~76, no. 3-4, pp. 539--562,
  2014.

\end{thebibliography}
}
\end{document}